\documentclass[10pt,twocolumn,letterpaper]{article}

\usepackage{wacv}
\usepackage{times}
\usepackage{epsfig}
\usepackage{graphicx}
\usepackage{amsmath}
\usepackage{amssymb}

\usepackage{breqn}
\usepackage{subcaption}

\usepackage{pifont}
\usepackage{booktabs}
\usepackage{array}
\usepackage{multirow}
\usepackage{float}
\usepackage{balance}
\usepackage[table]{xcolor}
\def\wacvPaperID{1333} 

\wacvfinalcopy 
\ifwacvfinal
\def\assignedStartPage{9876} 
\fi

\ifwacvfinal
\usepackage[breaklinks=true,bookmarks=false]{hyperref}
\else
\usepackage[pagebackref=true,breaklinks=true,colorlinks,bookmarks=false]{hyperref}
\fi

\ifwacvfinal
\else
\fi

\begin{document}

\title{CIT-GAN: Cyclic Image Translation Generative Adversarial Network With Application in Iris Presentation Attack Detection}

\author{Shivangi Yadav\\
Michigan State University\\
{\tt\small yadavshi@msu.edu}
\and
Arun Ross\\
Michigan State University\\
{\tt\small rossarun@msu.edu}
}

\maketitle

\begin{abstract}
  In this work, we propose a novel Cyclic Image Translation Generative Adversarial Network (CIT-GAN) for multi-domain style transfer. To facilitate this, we introduce a Styling Network that has the capability to learn style characteristics of each domain represented in the training dataset. The Styling Network helps the generator to drive the translation of images from a source domain to a reference domain and generate synthetic images with style characteristics of the reference domain. The learned style characteristics for each domain depend on both the style loss and domain classification loss. This induces variability in style characteristics within each domain. The proposed CIT-GAN is used in the context of iris presentation attack detection (PAD) to generate synthetic presentation attack (PA) samples for classes that are under-represented in the training set.  Evaluation using current state-of-the-art iris PAD methods demonstrates the efficacy of using such synthetically generated PA samples for training PAD methods. Further, the quality of the synthetically generated samples is evaluated using Frechet Inception Distance (FID) score. Results show that the quality of synthetic images generated by the proposed method is superior to that of other competing methods, including StarGan. 

\end{abstract}

\begin{figure}
\centering
 \includegraphics[width=0.7\linewidth]{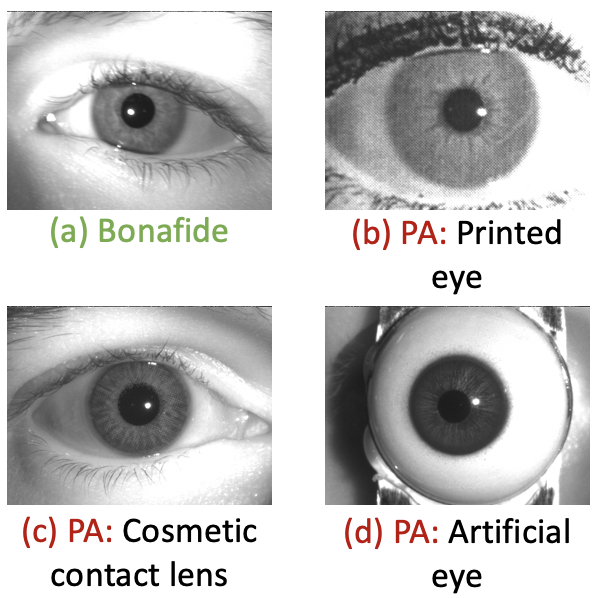}
 \caption{Examples of bonafide iris and presentation attack (PA) images: (a) Bonafide, (b) Printed eye, (c) Cosmetic contact lens and (d) Artificial eye} 
 \vspace{-4mm}
 \label{fig:Dataset}
\end{figure}

\section{Introduction}

The unique texture of the iris has made iris-based recognition systems desirable for human recognition in a number of applications \cite{jain2016}. However, these systems are increasingly facing threats from presentation attacks (PAs) where an adversary attempts to obfuscate their own identity, impersonate someone's identity, or create a virtual identity \cite{Ross2019}. Some commonly known iris presentation attacks (as shown in Figure \ref{fig:Dataset}) are: 

\begin{figure*}[t]
 \centering
 \includegraphics[width=0.90\linewidth]{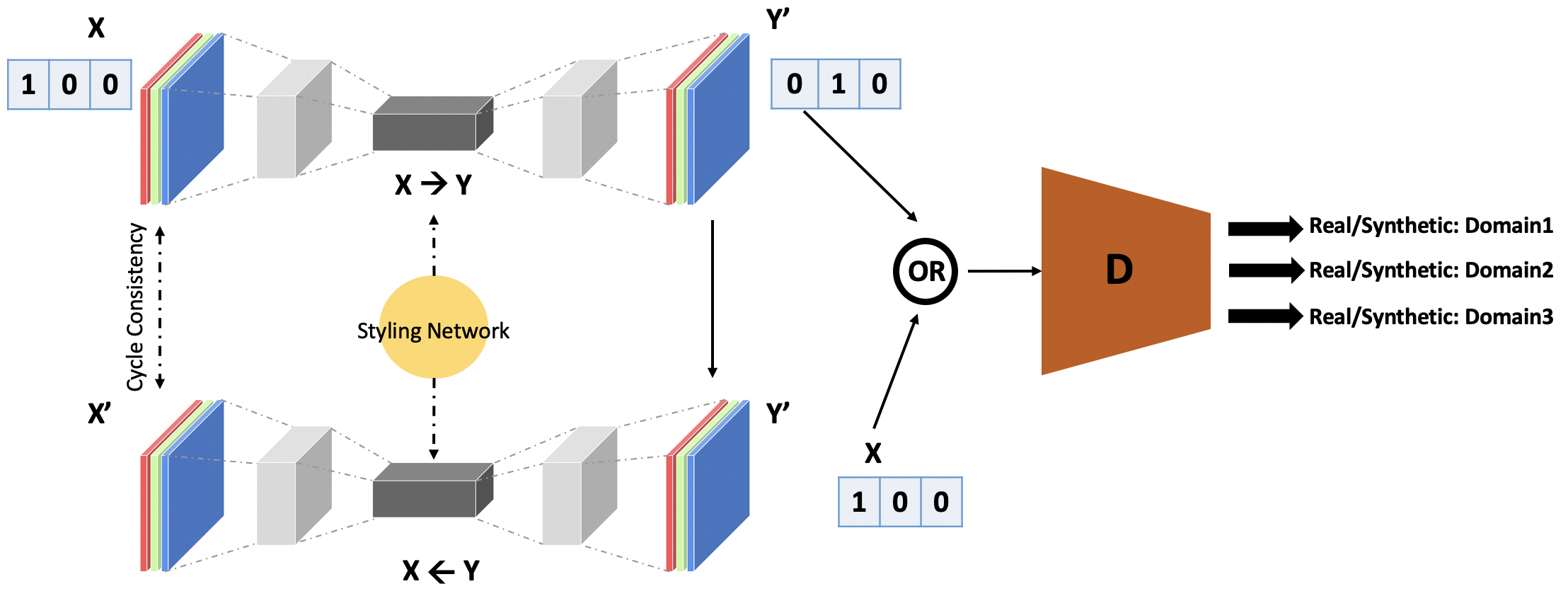}
 \caption{Schematic of the proposed Cyclic Image Translation Generative Adversarial Network (CIT-GAN). The proposed architecture has three important components: (a) Generator: Unlike a standard generator, this network takes as an image \ensuremath{X} from a domain and its label as input (represented as [1,0,0] in the figure) and outputs an image \ensuremath{Y'} with style similar to a reference image \ensuremath{Y} with domain label [0,1,0]. (b) Styling Network: The image-to-image translation to multiple domains is driven by a Styling Network that learns the style code for each given domain. (c) Discriminator: Unlike a standard discriminator, the discriminator in the proposed method has multiple branches each of which determines whether the input image is real or synthetic pertaining to that domain.}
 \vspace{-4mm}
 \label{fig:Propose}
\end{figure*}


\begin{figure}
 \centering
 \includegraphics[width=1\linewidth]{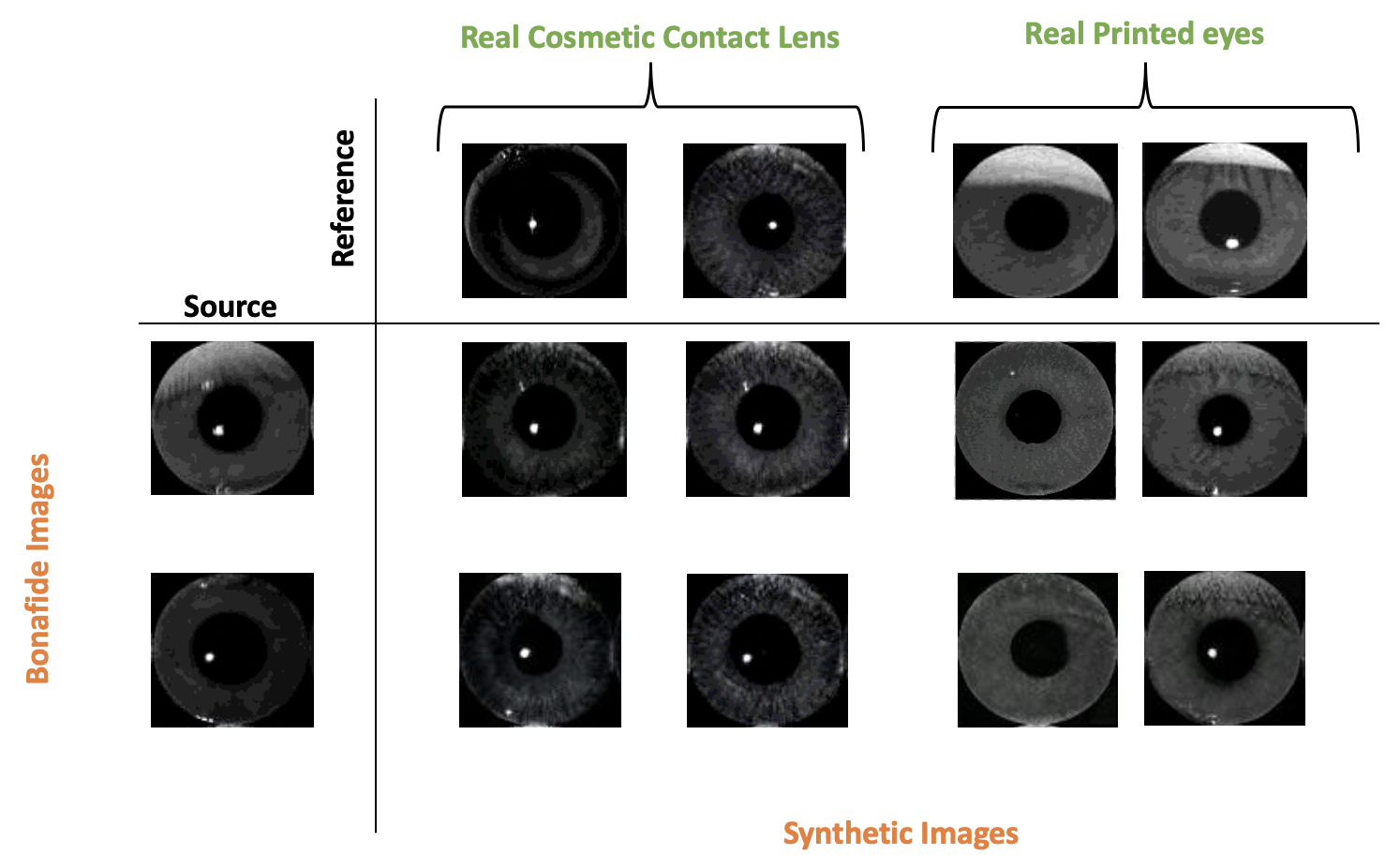}
 \caption{Examples of image translation from source to reference domain via CIT-GAN using domain specific styling vector obtained from the Styling Network.} 
 \vspace{-4mm}
 \label{fig:Translation}
\end{figure}


\begin{itemize}
    \item \textbf{Printed photo} \cite{bodade2011, gupta2014}: High-tech digital printers and scanners have made it possible to print good quality images of bonafide irides, which can be used to impersonate someone’s identity.
    \item \textbf{Artificial eyes} \cite{czajka2016iris}: Prosthetic or doll eyes are typically hand-crafted by professionals to look as similar to the bonafide irides as possible. Attackers can utilize such artifacts to obfuscate their true identity.
    \item \textbf{Cosmetic contact lens} \cite{daugman2003, doyle2013}: The term “cosmetic contact lens” refers to lenses that typically have texture over them and are tinted with some color. These patterns can obstruct the natural iris texture that is required to recognize a person. Therefore, they can be used to deceive recognition systems.
\end{itemize}

In addition, electronic displays \cite{hoffman2019}, cadaver eyes \cite{boyd2020} and holographic eye images \cite{pacut2006} may be used to launch a presentation attack. Grasping the threat posed by these attacks, researchers have been working on devising methods for iris presentation attack detection (PAD) that aim to distinguish between bonafide and PAs. In \cite{gupta2014, Raghav2015}, researchers used textural descriptors like Local Binary Pattern (LBP) and multi-scale binarized statistical image features (BSIF) to detect print attacks. Kohli et al. \cite{kohli2016} proposed a variation of LBP to obtain textual information from iris images that helps in detecting cosmetic contact lens. More recently, deep features from Convolutional Neural Networks (CNNs) have been used to detect multiple iris presentation attacks \cite{chen2018}\cite{hoffman2019}. Yadav et. al. \cite{yadav2020} utilized the Relativistic Discriminator from a RaSGAN as a one-class classifier for PA detection. These methods report high accuracy for iris PA detection, but their performance can be negatively impacted by the absence of a sufficient number of samples from different PA classes \cite{czajka2018}. Therefore, we can conclude that current iris PAD methods need a copious amount of training data corresponding to different PA classes and scenarios. Unfortunately, in the real world, such a dataset is hard to acquire.

\begin{figure*}[t]
 \centering
 \includegraphics[width=0.75\linewidth]{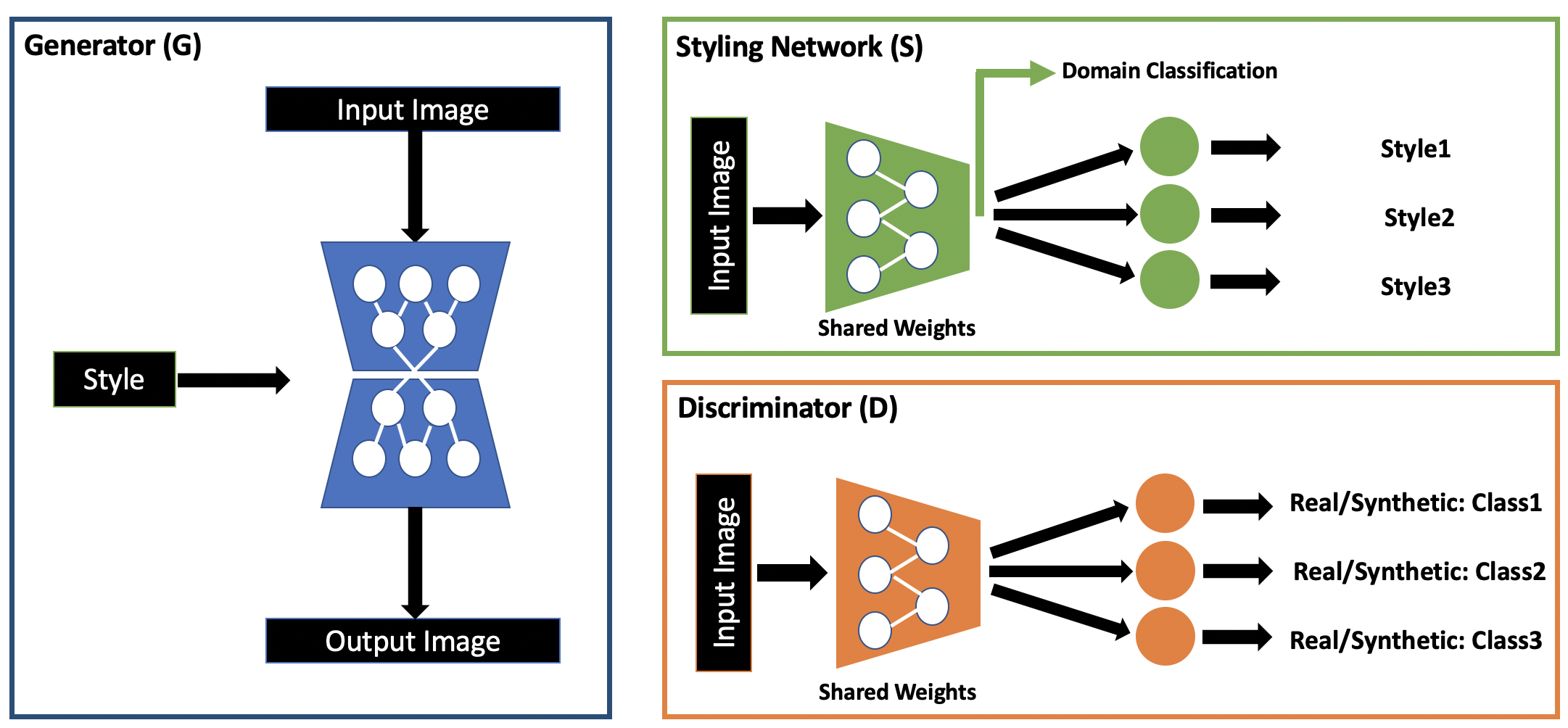}
 \caption{The components of proposed network Cyclic Image Translation Generative Adversarial Network. The Generator takes in an image as input and the image translation from one domain to another is driven by styling code from the Styling Network. Styling Network learns the style encoding or style characteristics of each domain, while domain classification loss ensures that the style encoding learnt by the network belongs to the correct domain. Similar to a standard discriminator, the proposed discriminator competes with the generator by deciding for each domain whether the input is real or synthetic.} 
 \vspace{-4mm}
 \label{fig:Arch}
\end{figure*}

\begin{figure*}[t]
 \centering
 \includegraphics[width=0.65\linewidth]{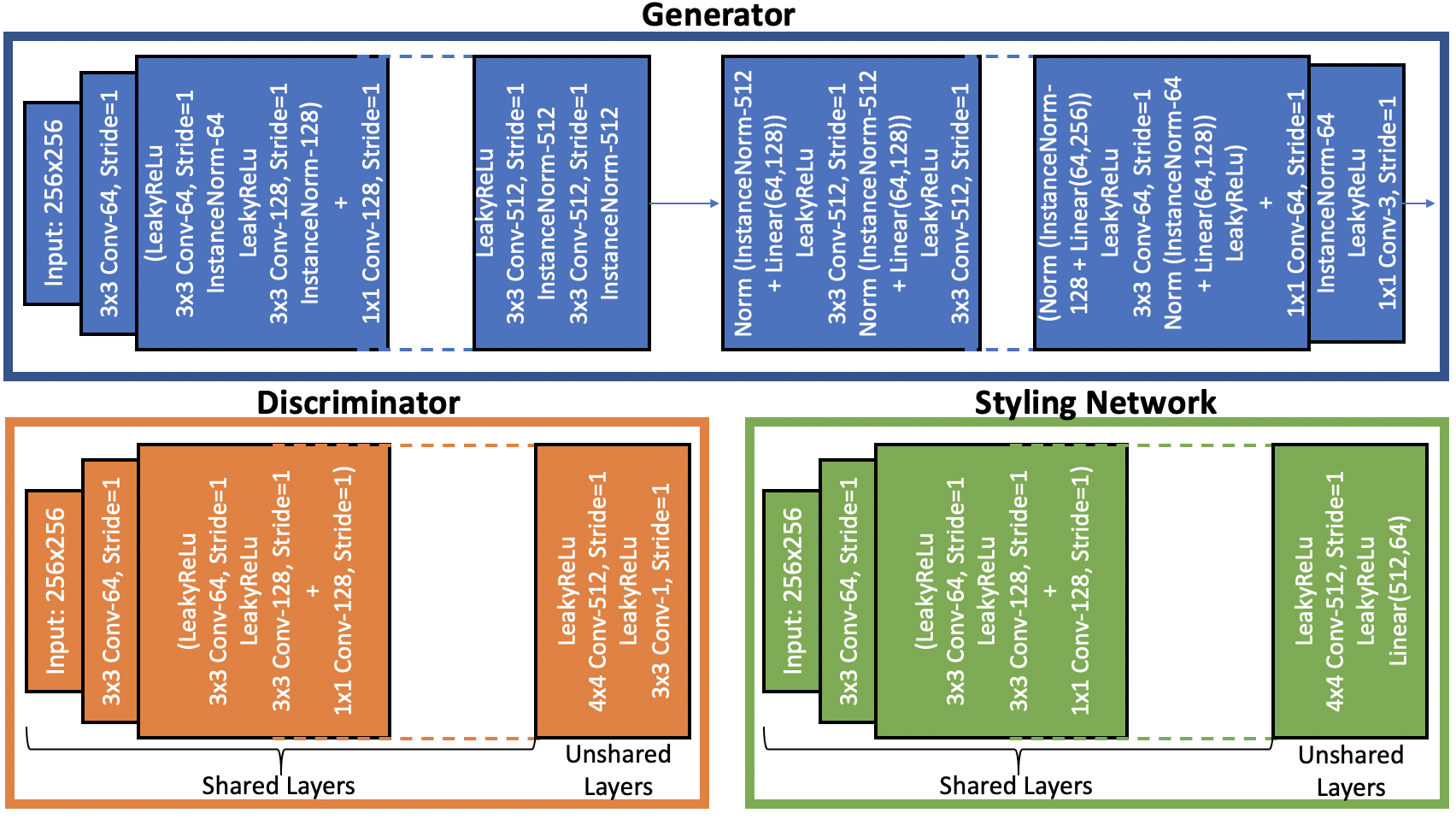}
 \caption{Details of the architecture of the CIT-GAN used for generating synthetic iris images.}
\label{fig:Model_Details}
\end{figure*}


With recent advances in the field of deep learning, researchers have proposed different methods based on Convolutional Autoencoders \cite{shrivastava2017,van2016} and Generative Adversarial Networks (GANs) \cite{goodfellow2014} for image-to-image style translation. Here, image-to-image translation refers to learning a mapping between different visual domain categories each of which has its own unique appearance and style. Gatys et al. \cite{gatys2015} proposed a neural architecture that could separate image content from style and then combine the two in different combinations to generate new natural looking styles. Their paper mainly focused on learning styles from well known artworks to generate new high quality and natural looking artwork. Karras et al. \cite{karras2019} introduced StyleGAN that uses a non-linear mapping function to embed a style driven latent code to generate images with different styles. However, since the input to the generator in StyleGAN is a noise vector, non-trivial efforts are required to transform image from one domain to another. Some researchers overcame this issue by enforcing an overlay between generator’s input and output for diversity in generated images using either marginal matching \cite{almahairi2018} or diversity regularization \cite{yang2019}. Others approached style transfer with the guidance of some reference images \cite{chang2018, cho2019}.

However, these methods are not scalable to more than two domains and often show instability in the presence of multiple domains \cite{choi2020}. Choi et al. \cite{choi2018, choi2020} proposed to solve this problem by using a unified GAN architecture called StarGAN for style transfer that can generate diverse images across multiple domains. StarGAN uses a single generator with a mapping and Styling Network to learn diverse mappings between all available domains. The Styling Network aims to learn style codes for all the domains, while the mapping network is used to produce random style codes from latent codes. On one hand, it enforces
the generator to learn diverse mappings across domains, but it also introduces the risk of generating images that are far from the original domains. Therefore, when introducing the mapping network, it becomes important to regularize the generator with diversity sensitive loss \cite{choi2020} that helps ensure that the generated data are diverse in nature.

\begin{figure*}[t]
 \centering
 \includegraphics[width=0.90\linewidth]{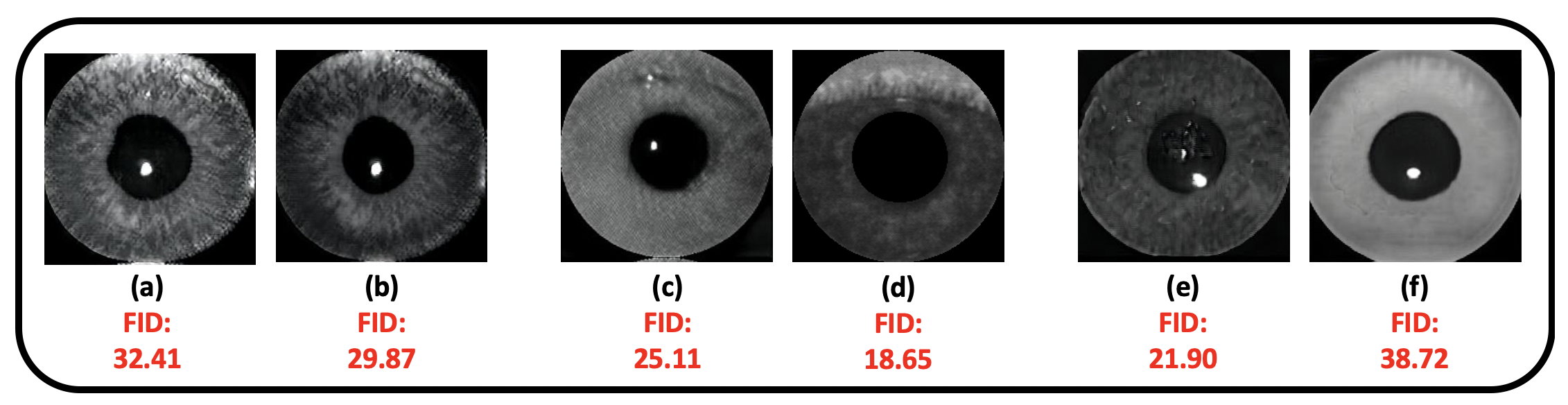}
 \caption{Given are some examples of synthetic samples generated using the proposed Cyclic Image Translation Generative Adversarial Network (CIT-GAN). (a)-(b) represent synthetic cosmetic contact lenses, (c)-(d) are two different types of synthetically generated print images (one with whole iris image and other with pupil cut-out), and (e)-(f) are synthetic artificial eyes with (f) representing a doll eye. } 
 \vspace{-4mm}
 \label{fig:Synthetic}
\end{figure*}


In this research, we propose a GAN architecture that uses a novel Styling Network to drive the translation of input image into multiple target domains. Here, the generator takes an image as input along with a domain label and then generates images with characteristics of the target domains. Apart from the domain label, the generative capability of the network is enhanced using a multi-task Styling Network that learns the style codes for multiple PA domains and helps the generator to synthesize images reflecting the style components of the target PA domains.\footnote{A {\em PA domain} refers to a specific PA category, e.g., printed image.} The domain-specific style characteristics learned using Styling Network depend on both style loss and domain classification. This ensures variability in style characteristics within each domain. Since there are multiple domains, the discriminator has multiple output branches to decide if a given image is real or synthetic for each of the domains (see Figure \ref{fig:Propose}). 

The primary contributions of this paper are as follows:
\begin{itemize}
    \item We propose a unified GAN architecture, referred to as Cyclic Image Translation Generative Adversarial Network (CIT-GAN), with novel Styling Network to generate good quality synthetic images for multiple domains. The quality of generated samples is evaluated using Fréchet Inception Distance (FID) Score \cite{heusel2017}.
    \item We demonstrate the usefulness of the synthetically generated data to train state-of-the-art iris PAD methods and improve their accuracy.
\end{itemize}


\section{Background}
Generative Adversarial Networks (GANs) have been used by researchers extensively for synthetic image generation \cite{goodfellow2014}, super-resolution \cite{cai2019}, anomaly detection \cite{zenati2018}, etc. due to their effectiveness in learning intricate features and texture of a given data distribution. Details of these networks are presented below.

\subsection{Standard Generative Adversarial Network (GAN)}
A Generative Adversarial Network \cite{goodfellow2014} consists of two components, a generator \ensuremath{G} and a discriminator \ensuremath{D}, that compete with each other. \ensuremath{G} takes as input a noise vector \ensuremath{z} and aims to generate good quality synthetic data that closely resembles the real data. On the other hand, \ensuremath{D} aims to distinguish between synthetically generated data and real data. This is denoted by a min-max objective function \ensuremath{O},

\begin{equation}\label{eq1}
\begin{split}
    \ensuremath{ \min\limits_{G}\max\limits_{D} O(D,G) = \mathbb{E}_{\boldsymbol{x_r} \sim \mathbb{P}}[log(D(\boldsymbol{x_r}))] \\
    + \mathbb{E}_{\boldsymbol{z} \sim \mathbb{M}}[log(1-D(G(\boldsymbol{z})))]}.
\end{split}
\end{equation}

Here, \ensuremath{\boldsymbol{x_r} \sim \mathbb{P}} represents the real data distribution and \ensuremath{\boldsymbol{z} \sim \mathbb{M}} represents the Gaussian noise distribution. \ensuremath{D(\boldsymbol{x})} outputs whether input \ensuremath{\boldsymbol{x}} belongs to the real distribution or not. The generator \ensuremath{G} takes an input \ensuremath{\boldsymbol{z}} to generate a synthetic image.

\subsection{Frechet Inception Distance (FID) Score}
Due to rapid advances in the field of DeepFakes, researchers have been studying different methods to evaluate the quality of synthetically generated data. Salimans et al. \cite{salimans2018} used a pre-trained inception-V3 to compare the marginal and conditional label distribution of synthetically generated data to compute the inception score. Higher the inception score, better the quality of the generated dataset. However, the inception score does not include the statistics of the real data distribution when computing the score. In \cite{heusel2017}, Heusel et al. exploited the statistics of real data and compared it with the statistics of the synthetically generated dataset to compute the Frechet Inception Distance (FID) score:
\begin{equation}\label{eq2}
\begin{split}
    \ensuremath{FID = \left \| \boldsymbol{\mu_r} - \boldsymbol{\mu_s} \right \|^2 + Tr(\boldsymbol{\Sigma_r} + \boldsymbol{\Sigma_s} - 2 \sqrt{\boldsymbol{\Sigma_r}\boldsymbol{\Sigma_s}}).}
\end{split}
\end{equation}
Here, \ensuremath{\boldsymbol{\mu_s}, \boldsymbol{\mu_r}, \boldsymbol{\Sigma_s} \text{ and } \boldsymbol{\Sigma_r}} are the statistics of the synthetic (\ensuremath{s}) and real (\ensuremath{r}) distributions. Since FID computes distance between the two distributions, the lower the FID score, better is the quality of generated dataset.


\begin{table*}[t]
\centering
\caption{Experiment-1: True Detection Rate (TDR in \%) at 0.1\%, 0.2\% and 1.0\% False Detection Rate (FDR) of existing iris PAD algorithms in baseline experiment (referred to as Experiment-1) when trained using imbalanced samples across different PA classes.}
\vspace{-2mm}
\label{tab:Exp1}
\scalebox{0.9}{%
\begin{tabular}{@{}lcccccc@{}}
\toprule
      & \multicolumn{1}{l}{\textbf{BSIF+SVM} \cite{Doyle2015}} & \multicolumn{1}{l}{\textbf{Fine-Tuned VGG-16} \cite{gatys2015}} & \multicolumn{1}{l}{\textbf{DESIST} \cite{kohli2016}} &
      \multicolumn{1}{l}{\textbf{Fine-Tuned AlexNet} \cite{alex2012}} &
      \multicolumn{1}{l}{\textbf{D-NetPAD} \cite{sharma20}}  \\ \midrule
TDR (@0.1\%)  & 3.32   & 85.25   & 4.25   & 86.10   & 87.94
\\
TDR (@0.2\%)  & 6.15   & 83.86   & 5.85   & 87.29   & 88.91 
\\
TDR (@1.0\%)    & 28.11  & 89.07   & 17.15  & 90.51   & 92.54
\\ \bottomrule
\end{tabular}}
\end{table*}

\begin{table*}[t]
\centering
\caption{Experiment-2: True Detection Rate (TDR in \%) at 0.1\%, 0.2\% and 1.0\% False Detection Rate (FDR) of existing iris PAD algorithms in Experiment-2 to evaluate the equivalence between real and synthetic PAs. When comparing with the performances in Table \ref{tab:Exp1}, it can be seen that substituting some of the real PAs in the training set with synthetically generated samples has a limited impact on the performance of PAD algorithms, especially at a FDR of 1\%.}
\vspace{-2mm}
\label{tab:Exp2}
\scalebox{0.9}{%
\begin{tabular}{@{}lcccccc@{}}
\toprule
      & \multicolumn{1}{l}{\textbf{BSIF+SVM} \cite{Doyle2015}} & \multicolumn{1}{l}{\textbf{Fine-Tuned VGG-16} \cite{gatys2015}} & \multicolumn{1}{l}{\textbf{DESIST} \cite{kohli2016}} &
      \multicolumn{1}{l}{\textbf{Fine-Tuned AlexNet} \cite{alex2012}} &
      \multicolumn{1}{l}{\textbf{D-NetPAD} \cite{sharma20}}  \\ \midrule
TDR (@0.1\%)  & 11.09    & 78.29   & 4.48   & 79.90   & 84.27  
\\
TDR (@0.2\%)  & 18.06    & 83.03   & 7.43   & 84.13   & 86.04
\\
TDR (@1.0\%)    & 29.38    & 87.39   & 17.43  & 89.66   & 89.55 
\\ \bottomrule
\end{tabular}}
\end{table*}

\section{Cyclic Image Translation Generative Adversarial Network (CIT-GAN)}
Let \ensuremath{ \boldsymbol{x} \sim \mathcal{X}} be an input image and \ensuremath{\boldsymbol{d} \sim \mathfrak{D}} be an arbitrary domain from domain space \ensuremath{\mathfrak{D}}. The proposed method aims to translate image \ensuremath{\boldsymbol{x}} to synthetic image \ensuremath{\boldsymbol{y}} with style characteristics of domain \ensuremath{\boldsymbol{d}}. This is achieved using a Styling Network \ensuremath{S} that is trained to {\em learn} domain specific style codes, and then train \ensuremath{G} to generate synthetic images with the given target style codes (see Figure \ref{fig:Translation}).

\subsection{Generative Adversarial Network}
Unlike standard GAN, the generative adversarial network in the proposed method has been updated to include domain level information. These changes are reflected in each component of the proposed architecture (as shown in \ref{fig:Arch}) :
\begin{itemize}
    \item \textbf{Generator}: For image-to-image translation between multiple domains, \ensuremath{G} takes an input image \ensuremath{\boldsymbol{x}} and translates it to an image \ensuremath{G(\boldsymbol{x},\boldsymbol{s})} with the desired style code \ensuremath{\boldsymbol{s}}. The style code \ensuremath{\boldsymbol{s}} is facilitated by Styling Network \ensuremath{S} and injected into G.
    \item \textbf{Discriminator}: Discriminator in the proposed architecture has multiple branches, where each branch \ensuremath{D_d} decides whether the input image \ensuremath{\boldsymbol{x}} is a real image in domain \ensuremath{\boldsymbol{d}} or a synthetic image. 
\end{itemize}

With the new objective for the generative adversarial network, the adversarial loss can be updated as,
\begin{equation}\label{eq3}
\begin{split}
    \ensuremath{ \mathcal{L}_{adv} = \mathbb{E}_{\boldsymbol{x,d}}
    [log(D_{\boldsymbol{d}}(\boldsymbol{x}))] \\
    + \mathbb{E}_{\boldsymbol{x,d'}}
    [log(1-D_{\boldsymbol{d'}}(G(\boldsymbol{x,s'})))]}.
\end{split}
\end{equation}

Here, \ensuremath{D_{\boldsymbol{d}}(x)} outputs a decision on image \ensuremath{\boldsymbol{x}} for domain branch \ensuremath{\boldsymbol{d}}. The Styling Network \ensuremath{S} takes an image \ensuremath{\boldsymbol{y}} from target domain \ensuremath{\boldsymbol{d'}} and outputs a style code \ensuremath{\boldsymbol{s'}}. \ensuremath{G(\boldsymbol{x,s'})} generates image \ensuremath{y'} with style characteristics of target domain \ensuremath{d'}.

\begin{figure}
 \centering
 \includegraphics[width=1\linewidth]{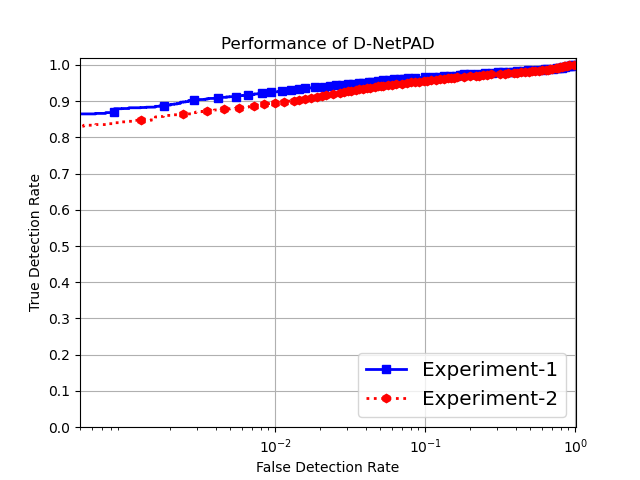}
 \caption{Comparing performance of D-NetPAD in Experiment-1 and Experiment-2 to evaluate the equivalence of synthetic PA samples in replacing real PA samples. D-NetPAD is one of the best performing PAD algorithms in iris liveness detection competition (LivDet-20 edition) \cite{das2020}.} 
 \vspace{-4mm}
 \label{fig:ROC1}
\end{figure}


\subsection{Styling Network}
Given an input image \ensuremath{\boldsymbol{x}} belonging to domain \ensuremath{\boldsymbol{d}}, the Styling Network \ensuremath{S} encodes the image into a style code \ensuremath{\boldsymbol{s}}. Similar to \ensuremath{D}, the Styling Network \ensuremath{S} is a multi-task network that learns the style code for an input image and injects the style code into \ensuremath{G} to generate images with the given style codes. This is achieved using \cite{choi2020},
\begin{equation}\label{eq4}
\begin{split}
    \ensuremath{ \mathcal{L}_{style} = \mathbb{E}_{\boldsymbol{x,d'}}
    [\| \boldsymbol{s'} - S_{\boldsymbol{d'}}(G(\boldsymbol{x,s'})) \| ]}.
\end{split}
\end{equation}
Here, \ensuremath{s' = S(y)} is the style code of reference image \ensuremath{\boldsymbol{y}} belonging to target domain \ensuremath{\boldsymbol{d'}}. This ensures that \ensuremath{G} generates images with the specified style code. However, poor quality synthetic data in the initial training iterations can affect the quality of the domain specific style codes learned by \ensuremath{S}. To avoid this, we introduce a domain classification loss \ensuremath{\mathcal{L}_{cls}} at the shared layer of \ensuremath{S} from soft-max layer (as shown in Figure \ref{fig:Arch}) to ensure that the learnt style code aligns with the correct domain. Further, this helps the Styling Network to learn style vectors (or feature characteristics) of varying samples from same domain.
\begin{equation}\label{eqcls}
    \ensuremath{ \mathcal{L}_{cls} = -logP(D= \boldsymbol{d} | X=\boldsymbol{x}).}
\end{equation}
Here, \ensuremath{\boldsymbol{d}} is the true domain of input \ensuremath{\boldsymbol{x}}.

\begin{table*}[t]
\centering
\caption{Experiment-3: True Detection Rate (TDR in \%) at 0.1\%, 0.2\% and 1.0\% False Detection Rate (FDR) of existing iris PAD algorithms in Experiment-3 to evaluate the efficacy of proposed method, CIT-GAN, in generating synthetic PA samples that captures the real PA distribution across various PA domains.}
\vspace{-2mm}
\label{tab:Exp3}
\scalebox{0.9}{%
\begin{tabular}{@{}lcccccc@{}}
\toprule
      & \multicolumn{1}{l}{\textbf{BSIF+SVM} \cite{Doyle2015}} & \multicolumn{1}{l}{\textbf{Fine-Tuned VGG-16} \cite{gatys2015}} & \multicolumn{1}{l}{\textbf{DESIST} \cite{kohli2016}} &
      \multicolumn{1}{l}{\textbf{Fine-Tuned AlexNet} \cite{alex2012}} &
      \multicolumn{1}{l}{\textbf{D-NetPAD} \cite{sharma20}}  \\ \midrule
TDR (@0.1\%)  & 7.57    & 76.12   & 2.31    & 78.89   & 82.26  
\\
TDR (@0.2\%)  & 10.51   & 80.98   & 4.70    & 82.66   & 84.54
\\
TDR (@1.0\%)    & 29.43   & 85.81   & 15.29   & 88.37   & 88.86
\\ \bottomrule
\end{tabular}}
\end{table*}

\begin{table*}[t]
\centering
\caption{Experiment-4: True Detection Rate (TDR in \%) at 0.1\%, 0.2\% and 1.0\% False Detection Rate (FDR) of existing iris PAD algorithms in Experiment-4. When comparing against the performance numbers in Table \ref{tab:Exp1}, it can be seen that training using balanced samples from each PA class/domain helps improve the performance of current iris PAD algorithms.}
\vspace{-2mm}
\label{tab:Exp4}
\scalebox{0.9}{%
\begin{tabular}{@{}lcccccc@{}}
\toprule
      & \multicolumn{1}{l}{\textbf{BSIF+SVM} \cite{Doyle2015}} & \multicolumn{1}{l}{\textbf{Fine-Tuned VGG-16} \cite{gatys2015}} & \multicolumn{1}{l}{\textbf{DESIST} \cite{kohli2016}} &
      \multicolumn{1}{l}{\textbf{Fine-Tuned AlexNet} \cite{alex2012}} &
      \multicolumn{1}{l}{\textbf{D-NetPAD} \cite{sharma20}}  \\ \midrule
TDR (@0.1\%)  & 14.39   & 69.40   & 2.59   & 79.26    & 90.38  
\\
TDR (@0.2\%)  & 22.91   & 75.34   & 5.38   & 82.80    & 94.19 
\\
TDR (@1.0\%)    & 51.11   & 91.60   & 21.41  & 92.70    & 97.89
\\ \bottomrule
\end{tabular}}
\end{table*}

\begin{figure}
 \centering
 \includegraphics[width=1\linewidth]{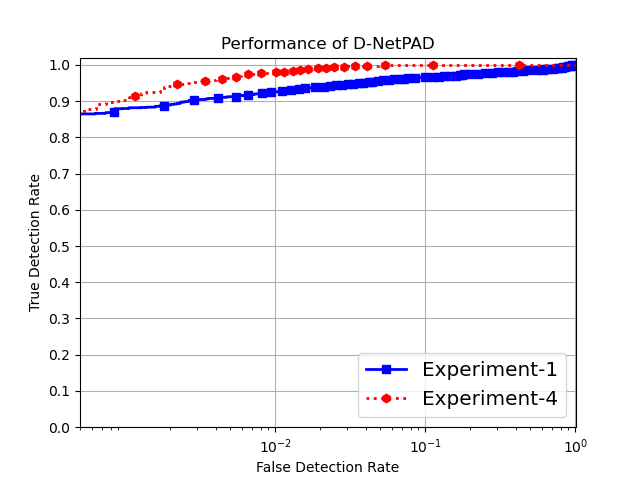}
 \caption{Comparing the performance of D-NetPAD in Experiment-1 and Experiment-4 to emphasize that the performance of current iris PAD methods are affected due to training with imbalanced samples from PA classes (Experiment-1). Improved performance is reported in Experiment-4 that utilizes synthetic samples for balanced training.}
 \vspace{-4mm}
 \label{fig:ROC2}
\end{figure}


\subsection{Cycle Consistency}
While translating images from the source domain to the domains depicted by the reference images, it is important to preserve some characteristics of the input images (such as geometry, pose and eye lashes in case of irides). This is achieved using the cycle consistency loss \cite{choi2018},
\begin{equation}\label{eq5}
\begin{split}
    \ensuremath{ \mathcal{L}_{cycle} = \mathbb{E}_{\boldsymbol{x,d,d'}}
    [\| \boldsymbol{x} - G(G(\boldsymbol{x,s'}),s) \| ]}.
\end{split}
\end{equation}
Here, \ensuremath{\boldsymbol{s} = S(x)} represents the style code of input image with domain \ensuremath{\boldsymbol{d}}, and \ensuremath{\boldsymbol{s'} = S(y)} is the style code of reference image in target domain \ensuremath{\boldsymbol{d'}}. This ensures that image \ensuremath{\boldsymbol{x}} with style \ensuremath{\boldsymbol{s}} can be reconstructed using synthetic image \ensuremath{G(\boldsymbol{x,s'})}.

Hence, the overall loss function for the proposed Cyclic Image Translation Generative Adversarial Network can be defined as:
\begin{equation}\label{eq6}
    \ensuremath{\mathcal{L}_{total} = \mathcal{L}_{adv} + \lambda_1 \mathcal{L}_{style} + \lambda_2 \mathcal{L}_{cls} + \lambda_3 \mathcal{L}_{cycle}.
    }
\end{equation}

Here, \ensuremath{\lambda_1, \lambda_2} and \ensuremath{\lambda_3} represent the hyperparameters for each loss term. 


\section{Dataset Used}
In this research, we utilized five different iris PA datasets, viz., Casia-iris-fake \cite{Sun2014}, Berc-iris-fake \cite{Lee2007}, NDCLD15 \cite{Doyle2015}, LivDet2017 \cite{Yambay2017} and MSU-IrisPA-01 \cite{Yadav2019} for training and testing different iris presentation attack detection (PAD) algorithms. These iris datasets contain bonafide images and images from different PA classes such as cosmetic contact lenses, printed eyes, artificial eyes and kindle-display attack (as shown in Figure \ref{fig:Dataset}). The images in these datasets are pre-processed and cropped to a size of 256x256 around the iris using the coordinates from a software called VeriEye.\footnote{\url{www.neurotechnology.com/verieye.html}} The images that were not properly processed by VeriEye were discarded from the datasets as this paper focuses primarily on image synthesis. This give us a total of 24,409 bonafide irides, 6,824 cosmetic contact lenses, 680 artificial eyes and 13,293 printed eyes.

\section{Image Quality}
The proposed architecture is trained using 6,450 bonafide images, 2,104 cosmetic contact lenses, 4,482 printed eyes and 276 artificial eyes randomly selected from the aforementioned datasets. The trained network is then utilized to generate synthetic PA samples. To achieve this, 6,000 bonafide images were utilized as source images. The source images are then translated to different PA classes using 2,000 printed eyes, 2,000 cosmetic contact lens and 276 artificial eyes as reference images. Using this approach, we generated 8,000 samples for each PA class. The generated samples from CIT-GAN obtained an average FID score of 32.79.

\begin{figure*}[t]
 \centering
 \includegraphics[width=1\linewidth]{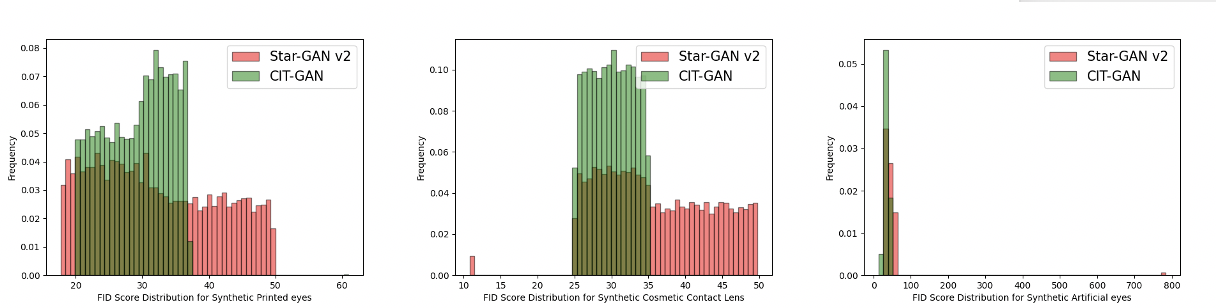}
 \caption{Comparing the FID score distributions of Star-GAN v2 \cite{choi2020} and CIT-GAN for each synthetically generated PA domain. A lower FID is better.} 
 \vspace{-4mm}
 \label{fig:Hist}
\end{figure*}

For comparison purposes, we used the same train and evaluation setup to generate synthetic samples using Star-GAN \cite{choi2018}, Star-GAN v2 \cite{choi2020} and Style-GAN \cite{karras2019}. As mentioned before, Style-GAN and Star-GAN are not well-equipped to handle multi-domain image translation. Therefore, they obtained a high average FID score of 86.69 and 44.76, respectively. On the other hand, Star-GAN v2 is equipped to handle multi-domains using a styling and mapping network. A trained Star-GAN v2 utilizes the mapping network to generate diverse style codes to diversify images. The synthetic iris PAs generated using this method were diverse in nature, but failed to capture the true characteristics of PAs like artificial eyes. Hence, the average FID score of the generated image using Star-GAN v2 was 38.81 - much lower than that of  Style-GAN and Star-GAN, but still a bit higher than CIT-GAN. This can also be seen in the FID score distribution in Figure \ref{fig:Hist} that compares the synthetically generated data using Star-GAN v2 with that of CIT-GAN.


\begin{figure}
 \centering
 \includegraphics[width=1\linewidth]{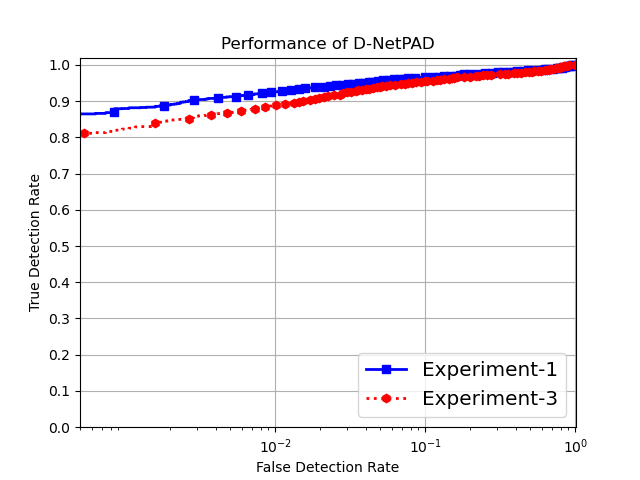}
 \caption{Comparing performance of D-NetPAD in Experiment-1 and Experiment-3 to evaluate the efficacy of the proposed method, CIT-GAN, in generating synthetic PA samples that represent the real PA distribution across various PA domains.}
 \vspace{-4mm}
 \label{fig:ROC3}
\end{figure}


\section{Experimental Setups}
In this section, we describe different experimental setups that are used to evaluate the quality and usefulness of synthetic PA samples generated using CIT-GAN. We evaluated the performance of different iris PAD methods viz., VGG-16 \cite{gatys2015}, BSIF \cite{Doyle2015}, DESIST \cite{kohli2016}, D-NetPAD \cite{sharma20} and AlexNet \cite{alex2012} under these different experimental setups for analysis purposes. Note that D-NetPAD is one of the best performing PAD algorithms in the iris liveness detection competition (LivDet-20 edition) \cite{das2020}.

\subsection{Experiment-1}
This is the baseline experiment that demonstrates the performance of the current iris PAD methods on the previously mentioned datasets with imbalanced samples across different PA classes. The PAD methods are trained using 14,970 bonafide samples and 10,306 PA samples consisting of 276 artificial eyes, 4,014 cosmetic contact lenses and 6,016 printed eyes. The test set consists of 9,439 bonafide samples and 9,896 PA samples corresponding to 404 artificial eyes, 2,720 cosmetic contact lenses and 6,772 printed eyes.

\subsection{Experiment-2}
In this experiment, our aim is to evaluate the equivalence between real PAs and synthetically generated PAs. Therefore, the iris PAD methods are trained using 14,970 bonafide samples and PAs consisting of both real PA images and synthetic PA images. The real PA dataset has 138 artificial eyes, 2,007 cosmetic contact lenses and 3,008 printed eyes. The synthetic PA dataset is generated using the remainder of the real PA dataset as reference images (i.e., 138 artificial eyes, 2,007 cosmetic contact lenses and 3,008 printed eyes) in order to capture their style characteristics in the generated dataset. As before, the test set consists of 9,439 bonafide and 9,896 PA samples corresponding to 404 artificial eyes, 2,720 cosmetic contact lenses and 6,772 printed eyes.

\subsection{Experiment-3}
This experiment aims to evaluate the efficacy of the proposed method, CIT-GAN, in generating synthetic PA samples that represent the real PA distribution across various PA domains. Here, the iris PAD methods are trained using 14,970 bonafide samples and synthetically generated 276 artificial eyes, 4,014 cosmetic contact lenses and 6,016 printed eyes. The test set consists of 9,439 bonafide and 9,896 PA samples corresponding to 404 artificial eyes, 2,720 cosmetic contact lenses and 6,772 printed eyes.

\subsection{Experiment-4}
As mentioned in Experiment-1, current iris PAD methods are trained and tested on imbalanced samples from PA classes thereby affecting their accuracy. To overcome this, we train the iris PAD methods using 14,970 bonafide samples and a balanced set of 15,000 PA samples corresponding to 276 artificial eyes, 4,014 cosmetic contact lenses and 5000 printed eyes that are real; and 4,724 artificial eyes and 986 cosmetic contact lenses that are synthetic. This balances the number of samples across PA classes. The testing was done on 9,439 bonafide samples and 9,896 PA samples consisting of 404 artificial eyes, 2,720 cosmetic contact lenses and 6,772 printed eyes.


\section{Results and Analysis}
In this section, we discuss the results obtained for the four different experiments described in the previous section. Experiment-1 is the baseline experiment that evaluates the performance of various iris PAD methods. The training set for this experiment contains 14,970 bonafide and 10,306 PA samples consisting of 276 artificial eyes, 4,014 cosmetic contact lenses and 6,016 printed eyes. Due to imbalance in the number of samples across various PA domains, the performance of the PAD methods is affected. This becomes apparent when comparing the results of Experiment-1 with that of Experiment-4 where PAD methods are trained using 9,439 bonafide samples and a balanced number of PA samples (i.e., 5,000 samples from each PA domain) containing both real and synthetic PAs. As seen from the results in Table \ref{tab:Exp1} and Table \ref{tab:Exp4}, performance for each PAD method improves in Experiment-4. For example, in the case of D-NetPAD, the TDR at a 1\% FDR improved from 92.54\% in Experiment-1 to 97.89\% in Experiment-4 (as shown in Figure \ref{fig:ROC2}). A huge increase in performance was also noticed
for BSIF+SVM where TDR improved from 28.11\% in Experiment-1 to 51.11\% in Experiment-4, at a FDR of 1\%.

In addition, the equivalence of synthetically generated PA samples and real PA samples was established using Experiment-2 and Experiment-3. In Experiment-2, some of the real PA samples in the training set were replaced with synthetically generated PAs. Comparing the performance in Table \ref{tab:Exp1} and Table \ref{tab:Exp2}, a very slight difference in PAD performance is observed (see Figure \ref{fig:ROC1}). Similarly, in Experiment-3, where all the real PAs are replaced with synthetically generated PA samples, only a slight decrease in performance was seen for the PAD methods (as shown in Figure \ref{fig:ROC3}) signifying underlying similarities between real and synthetically generated data.

\section{Conclusion and Future Work}
In this research work, we designed an image-to-image translation method (CIT-GAN) that can synthetically generate images across multiple iris presentation attack domains, i.e., multiple types of iris PAs. The results obtained in the previous sections show the equivalence of synthetically generated PA samples and real PA samples. Furthermore, the results in Table \ref{tab:Exp1} and Table \ref{tab:Exp4} demonstrate that the performance of the iris PAD methods can be improved by adding synthetically generated data to different PA classes for balanced training.

We would like to extend this work to generate PA samples with different variations and multi-domain style characteristics such as "printed" cosmetic contact lenses, etc. This will allow current iris PAD methods to generalize over different styles of PAs for enhanced PA detection.


\section{Acknowledgment}
{\small This research is based upon work supported in part by the Office of the Director of National Intelligence (ODNI), Intelligence Advanced Research Projects Activity (IARPA), via IARPA R\&D Contract No. 2017 - 17020200004. The views and conclusions contained herein are those of the authors and should not be interpreted as necessarily representing the official policies, either expressed or implied, of ODNI, IARPA, or the U.S. Government. The U.S. Government is authorized to reproduce and distribute reprints for governmental purposes notwithstanding any copyright annotation therein.
}

{\small
\balance
\bibliographystyle{ieee_fullname}
\bibliography{egbib}

\begin{thebibliography}{10}\itemsep=-1pt

\bibitem{almahairi2018}
Amjad Almahairi, Sai Rajeswar, Alessandro Sordoni, Philip Bachman, and Aaron
  Courville.
\newblock Augmented {cycleGAN}: Learning many-to-many mappings from unpaired
  data.
\newblock {\em arXiv preprint arXiv:1802.10151}, 2018.

\bibitem{bodade2011}
Rajesh Bodade and Sanjay Talbar.
\newblock Fake iris detection: A holistic approach.
\newblock {\em International Journal of Computer Applications}, 19(2):1--7,
  2011.

\bibitem{boyd2020}
Aidan Boyd, Shivangi Yadav, Thomas Swearingen, Andrey Kuehlkamp, Mateusz
  Trokielewicz, Eric Benjamin, Piotr Maciejewicz, Dennis Chute, Arun Ross,
  Patrick Flynn, et~al.
\newblock {Post-Mortem Iris Recognition—A Survey and Assessment of the State
  of the Art}.
\newblock {\em IEEE Access}, 8:136570--136593, 2020.

\bibitem{cai2019}
Jiancheng Cai, Han Hu, Shiguang Shan, and Xilin Chen.
\newblock {Fcsr-GAN}: End-to-end learning for joint face completion and
  super-resolution.
\newblock In {\em 14th IEEE International Conference on Automatic Face \&
  Gesture Recognition (FG 2019)}, pages 1--8. IEEE, 2019.

\bibitem{chang2018}
Huiwen Chang, Jingwan Lu, Fisher Yu, and Adam Finkelstein.
\newblock Paired {cycleGAN}: Asymmetric style transfer for applying and
  removing makeup.
\newblock In {\em Proceedings of the IEEE Conference on Computer Vision and
  Pattern Recognition}, pages 40--48, 2018.

\bibitem{chen2018}
Cunjian Chen and Arun Ross.
\newblock {A multi-task convolutional neural network for joint iris detection
  and presentation attack detection}.
\newblock In {\em IEEE Winter Applications of Computer Vision Workshops
  (WACVW)}, pages 44--51, 2018.

\bibitem{cho2019}
Wonwoong Cho, Sungha Choi, David~Keetae Park, Inkyu Shin, and Jaegul Choo.
\newblock Image-to-image translation via group-wise deep whitening-and-coloring
  transformation.
\newblock In {\em Proceedings of the IEEE Conference on Computer Vision and
  Pattern Recognition}, pages 10639--10647, 2019.

\bibitem{choi2018}
Yunjey Choi, Minje Choi, Munyoung Kim, Jung-Woo Ha, Sunghun Kim, and Jaegul
  Choo.
\newblock {StarGAN}: Unified generative adversarial networks for multi-domain
  image-to-image translation.
\newblock In {\em Proceedings of the IEEE Conference on Computer Vision and
  Pattern Recognition}, pages 8789--8797, 2018.

\bibitem{choi2020}
Yunjey Choi, Youngjung Uh, Jaejun Yoo, and Jung-Woo Ha.
\newblock {StarGAN} v2: Diverse image synthesis for multiple domains.
\newblock In {\em Proceedings of the IEEE/CVF Conference on Computer Vision and
  Pattern Recognition}, pages 8188--8197, 2020.

\bibitem{czajka2016iris}
Adam Czajka.
\newblock Iris liveness detection by modeling dynamic pupil features.
\newblock In {\em Handbook of Iris Recognition}, pages 439--467. Springer,
  2016.

\bibitem{czajka2018}
Adam Czajka and Kevin~W Bowyer.
\newblock Presentation attack detection for iris recognition: An assessment of
  the state-of-the-art.
\newblock {\em ACM Computing Surveys (CSUR)}, 51(4):1--35, 2018.

\bibitem{das2020}
Priyanka Das, Joseph McGrath, Zhaoyuan Fang, Aidan Boyd, Ganghee Jang, Amir
  Mohammadi, Sandip Purnapatra, David Yambay, S{\'e}bastien Marcel, Mateusz
  Trokielewicz, et~al.
\newblock Iris liveness detection competition ({LivDet-Iris})--the 2020
  edition.
\newblock {\em In IEEE International Joint Conference on Biometrics (IJCB)},
  2020.

\bibitem{daugman2003}
John Daugman.
\newblock Demodulation by complex-valued wavelets for stochastic pattern
  recognition.
\newblock {\em International Journal of Wavelets, Multiresolution and
  Information Processing}, 1(01):1--17, 2003.

\bibitem{Doyle2015}
James~S Doyle and Kevin~W Bowyer.
\newblock Robust detection of textured contact lenses in iris recognition using
  {BSIF}.
\newblock {\em IEEE Access}, 3:1672--1683, 2015.

\bibitem{doyle2013}
James~S Doyle, Patrick~J Flynn, and Kevin~W Bowyer.
\newblock Automated classification of contact lens type in iris images.
\newblock In {\em International Conference on Biometrics (ICB)}, pages 1--6.
  IEEE, 2013.

\bibitem{gatys2015}
Leon~A Gatys, Alexander~S Ecker, and Matthias Bethge.
\newblock A neural algorithm of artistic style.
\newblock {\em arXiv preprint arXiv:1508.06576}, 2015.

\bibitem{goodfellow2014}
Ian Goodfellow, Jean Pouget-Abadie, Mehdi Mirza, Bing Xu, David Warde-Farley,
  Sherjil Ozair, Aaron Courville, and Yoshua Bengio.
\newblock Generative {Adversarial} {Nets}.
\newblock In {\em Advances in {Neural} {Information} {Processing} {Systems}
  (NIPS)}, pages 2672--2680, 2014.

\bibitem{gupta2014}
Priyanshu Gupta, Shipra Behera, Mayank Vatsa, and Richa Singh.
\newblock On iris spoofing using print attack.
\newblock In {\em 22nd International Conference on Pattern Recognition}, pages
  1681--1686. IEEE, 2014.

\bibitem{heusel2017}
Martin Heusel, Hubert Ramsauer, Thomas Unterthiner, Bernhard Nessler, and Sepp
  Hochreiter.
\newblock {GANs} trained by a two time-scale update rule converge to a local
  nash equilibrium.
\newblock In {\em Advances in neural information processing systems}, pages
  6626--6637, 2017.

\bibitem{hoffman2019}
Steven Hoffman, Renu Sharma, and Arun Ross.
\newblock {Iris + Ocular: Generalized Iris Presentation Attack Detection Using
  Multiple Convolutional Neural Networks}.
\newblock In {\em IAPR International Conference on Biometrics (ICB)}, 2019.

\bibitem{jain2016}
Anil~K Jain, Karthik Nandakumar, and Arun Ross.
\newblock 50 years of biometric research: Accomplishments, challenges, and
  opportunities.
\newblock {\em Pattern Recognition Letters}, 79:80--105, 2016.

\bibitem{karras2019}
Tero Karras, Samuli Laine, and Timo Aila.
\newblock A style-based generator architecture for generative adversarial
  networks.
\newblock In {\em Proceedings of the IEEE conference on computer vision and
  pattern recognition}, pages 4401--4410, 2019.

\bibitem{kohli2016}
Naman Kohli, Daksha Yadav, Mayank Vatsa, Richa Singh, and Afzel Noore.
\newblock {Detecting medley of iris spoofing attacks using DESIST}.
\newblock In {\em IEEE International Conference on Biometrics Theory,
  Applications and Systems (BTAS)}, pages 1--6, 2016.

\bibitem{alex2012}
Alex Krizhevsky, Ilya Sutskever, and Geoffrey~E Hinton.
\newblock Imagenet classification with deep convolutional neural networks.
\newblock In {\em Advances in Neural Information Processing Systems}, pages
  1097--1105, 2012.

\bibitem{Lee2007}
Sung~Joo Lee, Kang~Ryoung Park, Youn~Joo Lee, Kwanghyuk Bae, and Jai~Hie Kim.
\newblock Multifeature-based fake iris detection method.
\newblock {\em Optical Engineering}, 46(12):1--10, 2007.

\bibitem{pacut2006}
Andrzej Pacut and Adam Czajka.
\newblock Aliveness detection for iris biometrics.
\newblock In {\em Proceedings of 40th Annual International Carnahan Conference
  on Security Technology}, pages 122--129. IEEE, 2006.

\bibitem{Raghav2015}
R. Raghavendra and C. Busch.
\newblock Presentation attack detection algorithm for face and iris biometrics.
\newblock In {\em European Signal Processing Conference (EUSIPCO)}, pages
  1387--1391, 2014.

\bibitem{Ross2019}
Arun Ross, Sudipta Banerjee, Cunjian Chen, Anurag Chowdhury, Vahid Mirjalili,
  Renu Sharma, Thomas Swearingen, and Shivangi Yadav.
\newblock {Some Research Problems in Biometrics: The Future Beckons}.
\newblock In {\em IAPR International Conference on Biometrics (ICB)}, 2019.

\bibitem{salimans2018}
Tim Salimans, Han Zhang, Alec Radford, and Dimitris Metaxas.
\newblock Improving {GANs} using optimal transport.
\newblock {\em arXiv preprint arXiv:1803.05573}, 2018.

\bibitem{sharma20}
Renu Sharma and Arun Ross.
\newblock {D-NetPAD: An Explainable and Interpretable Iris Presentation Attack
  Detector}.
\newblock In {\em IEEE International Joint Conference on Biometrics (IJCB)},
  2020.

\bibitem{shrivastava2017}
Ashish Shrivastava, Tomas Pfister, Oncel Tuzel, Joshua Susskind, Wenda Wang,
  and Russell Webb.
\newblock Learning from simulated and unsupervised images through adversarial
  training.
\newblock In {\em Proceedings of the IEEE Conference on Computer Vision and
  Pattern Recognition (CVPR)}, pages 2107--2116, 2017.

\bibitem{Sun2014}
Z. Sun, H. Zhang, T. Tan, and J. Wang.
\newblock Iris image classification based on hierarchical visual codebook.
\newblock {\em IEEE Transactions on Pattern Analysis and Machine Intelligence},
  36(6):1120--1133, 2014.

\bibitem{van2016}
Aaron Van~den Oord, Nal Kalchbrenner, Lasse Espeholt, Oriol Vinyals, Alex
  Graves, et~al.
\newblock Conditional image generation with {CNN} decoders.
\newblock In {\em Advances in Neural Information Processing Systems (NIPS)},
  pages 4790--4798, 2016.

\bibitem{Yadav2019}
Shivangi Yadav, Cunjian Chen, and Arun Ross.
\newblock {Synthesizing Iris Images using RaSGAN with Application in
  Presentation Attack Detection}.
\newblock In {\em IEEE Conference on Computer Vision and Pattern Recognition
  Workshops (CVPRW)}, 2019.

\bibitem{yadav2020}
Shivangi Yadav, Cunjian Chen, and Arun Ross.
\newblock {Relativistic Discriminator: A One-Class Classifier for Generalized
  Iris Presentation Attack Detection}.
\newblock In {\em IEEE Winter Conference on Applications of Computer Vision},
  pages 2635--2644, 2020.

\bibitem{Yambay2017}
David Yambay, Benedict Becker, Naman Kohli, Daksha Yadav, Adam Czajka, Kevin~W
  Bowyer, Stephanie Schuckers, Richa Singh, Mayank Vatsa, Afzel Noore, et~al.
\newblock {LivDet Iris 2017 - }{Iris Liveness Detection Competition}.
\newblock In {\em IEEE International Joint Conference on Biometrics (IJCB)},
  pages 733--741, 2017.

\bibitem{yang2019}
Dingdong Yang, Seunghoon Hong, Yunseok Jang, Tianchen Zhao, and Honglak Lee.
\newblock Diversity-sensitive conditional generative adversarial networks.
\newblock {\em arXiv preprint arXiv:1901.09024}, 2019.

\bibitem{zenati2018}
Houssam Zenati, Chuan~Sheng Foo, Bruno Lecouat, Gaurav Manek, and
  Vijay~Ramaseshan Chandrasekhar.
\newblock Efficient {GAN-based} anomaly detection.
\newblock {\em arXiv preprint arXiv:1802.06222}, 2018.

\end{thebibliography}
}

\end{document}